# Run-Time Monitoring of Machine Learning for Robotic Perception: A Survey of Emerging Trends

**Quazi Marufur Rahman**  **Peter Corke, Fellow, IEEE**  **Feras Dayoub**

## Abstract

As deep learning continues to dominate all state-of-the-art computer vision tasks, it is increasingly becoming an essential building block for robotic perception. This raises important questions concerning the safety and reliability of learning-based perception systems. There is an established field that studies safety certification and convergence guarantees of complex software systems at design-time. However, the unknown future deployment environments of an autonomous system and the complexity of learning-based perception make the generalization of design-time verification to run-time problematic. In the face of this challenge, more attention is starting to focus on run-time monitoring of performance and reliability of perception systems with several trends emerging in the literature. This paper attempts to identify these trends and summarise the various approaches to the topic.

*Keywords* Machine learning, performance evaluation, reliability, robot learning.

## 1 Introduction

Deep Neural Networks (DNNs) show impressive results on many computer vision tasks such as image classification [71], object detection [42], depth estimation [77] and semantic segmentation [5]. This has led to their increased use for the perception pipeline of robotic and autonomous systems such as driverless cars, service, agricultural and field robots [27, 64, 70, 10]. However, a growing body of research is showing that state-of-the-art DNNs suffer a drop in performance when tested on data that differs from their training and testing sets [59, 50, 67]. This fact is of particular importance for deep learning based robotic perception since a robot may experience a wide range of environmental conditions that were not represented in the training data. This can lead to unexpected perception failures which pose an unacceptable safety risk. Without the ability to assess the reliability of the deep learning based components of the robotic system at run-time, the whole system's safety must be questioned.

The core of the problem is that, deep learning models are currently developed using a large dataset, split into training and test samples. As a result, the samples in the two sets are generated from the same distribution. In addition to that, most DNNs are trained with a closed-world assumption where all inputs are assumed to belong to one of a set of known categories. However, this is not always the case in the perception pipeline of a robot. The models' input data might come from unseen or different distributions than the training and testing sets due to environmental variables and novel contents that were not represented during design-time (i.e., the development and training stage). If this critical issue is not addressed, we can not generalize the model's performance on the test set to predict the performance during actual run-time (i.e., post-deployment on the robot) in a meaningful way.

Although there is an increasing interest in the area of safety certification and convergence guarantees for deep learning models at design-time (see [35] for a comprehensive overview), most of the current methods do not scale to large deep neural networks that are typical of what is used for robotic perception. In addition to this scale issue, there are the open-world deployment conditions that some mobile robots operate under (e.g. autonomous vehicles and

This work was supported in part by the Australian Research Council (ARC) Centre of Excellence for Robotic Vision under Grant CE140100016, and in part by the QUT Centre for Robotics.

Authors are from The ARC Centre of Excellence for Robotic Vision and The Centre for Robotics, Queensland University of Technology, Brisbane, QLD 4000, Australia

Corresponding author: Quazi Marufur Rahman (quazi.rahman@qut.edu.au)

field robots) that make the achievement of safety certification and convergence guarantees during design-time extra challenging. Consequently, the focus is shifting towards verification, validation and monitoring at run-time. Run-time monitoring checks a mobile robot's performance at its deployment phase, where ground-truth labels are not available. This monitoring is critically important for mobile robots' safety, and reliability as performance monitoring can work as a trigger to hand-over control to a less-capable-but-safe system or a human operator or shift to a fail-safe mode. To this end, this paper identifies and discusses emerging research trends that address the run-time performance monitoring of the learning-based components in autonomous robotic systems.

The paper is organized as follows: Section 2 presents our categorization of the reviewed papers based on *how* they perform the run-time monitoring. In Section 3, we revisit the same papers and re-categorize them based on *where* in the perception pipeline they perform the monitoring (i.e., whether at the input stage or in the target model itself or at its output stage or a combination of the above). Finally, we conclude with general observations in Section 4.

## 2 Run-Time Monitoring

Although run-time monitoring of machine learning for robotic perceptions is an emerging research topic, we can identify several trends in the literature based on their approach of detecting or predicting run-time failures. The first trend includes techniques that utilize past examples of failures or predicting the quality of the output based on the similarity of context or the place of operation to previous experiences. The second trend includes methods that detect inconsistencies in the perception output, either over a stream of input data or input from different sensors or outputs from different models. The third trend is based on confidence learning and uncertainty estimation, where perception modules express their own confidence in their output.

### 2.1 Monitoring Based on past experiences

This section will discuss the literature that focuses on monitoring perception system performance based on previous experience. We can categorize this literature into two groups. The first group monitors perception performance using past examples of success and failures, and the second group uses the experience from the same workspace or context for performance monitoring.

#### 2.1.1 Past Examples of Success and Failures

Generally, performance monitoring using examples of failure depends on an auxiliary network to predict the base network's failure. The base network can be responsible for any specific task – image classification, segmentation or object detection. The auxiliary network is trained using both positive and negative samples where the base network performed its particular task with expected accuracy. During the deployment phase, the auxiliary network operates along with the base network and predicts the base network's success or failure for performing the specific task.

The idea of using past examples of failures for the training of a self-evaluation system that detects perception failures at run-time has roots before the breakthrough of deep learning. An early example is the work by Jammalamadak *et al*. [38] where *evaluator algorithms* were proposed to predict the accuracy of a human pose estimation algorithm. They introduced the idea of self-evaluation and framed that as a binary classification task that uses additional features extracted from the target model's output. The binary classifier is trained as the evaluator, using examples of failures on the target model's training set. During inference, a threshold on the evaluator's quality is used to determine the human pose estimator's successes and failures.

Following similar approach to [38], Zhang *et al*. [76] proposed *alert* – a generalized warning framework to detect the failure of any vision system. *Alert* uses multiple generic hand-crafted image features to predict the accuracy of the vision system on that particular input image. They also introduced two new metrices – accuracy of vision system versus declaration rate, and risk-averse metric to evaluate the proposed performance prediction algorithm. Experimental results have shown *alert* is effective in predicting the failure of image segmentation, 3D layout estimation, image memorability and attributes-based scene and object recognition tasks. Daftry *et al*. [15] applied the *alert* framework to predict perception failure of an autonomous navigation task. They trained the *alert* system to predict Micro-Air Vehicle (MAV) navigation failure from the input image and corresponding optical flow. Here, *alert* is designed using spatiotemporal convolutional neural network for feature extraction and Support Vector Machine to identify cases where MAV will fail to navigate safely. Using a similar framework, Saxena *et al*. [65] trained the vision system of an autonomous quadrotor to identify navigational failure and response accordingly to avoid the consequences.

Joining the trend of using a separate system to monitor and predict a target model's failure, Mohseni *et al*. [48] proposed an approach to train a student model to predict the target model's error for input instances based on a saliency map extracted from the input images. The failure predictor is trained on examples of steering angle prediction errors of the target model for frames from the training set. In [69], a secondary model is trained using the softmax probabilities outputs of a target model to predict if its predictions are correct or not, therefore estimating the true inference accuracy on new and unseen data. The accuracy monitoring model needs to be pre-trained using data relevant to the target domain.

Most recently, Rahman *et al*. [56] addressed the problem of run-time performance monitoring of object detection onboard a mobile robot. They focused on the performance difference between the training and testing environment. They emphasized tracking the performance at run-time for the safety and reliability of object detection system during deployment. This work proposed a cascaded neural network to monitor performance by predicting the mean-average-precision metric over a sliding window of input images. In related work, [57] used an object detector's internal features to predict if the mean-average-precision for a particular image will be higher or lower than a predefined threshold. Identifying the false-negative object has been used by [55] as a means of run-time performance monitoring of object detection. In this work, they exploited features from specific feature map locations to identify potential false-negative objects. In a similar context, Schubert *et al*. [66] proposed a meta-classifier to discriminate between true-positive and false-positive, and performed meta-regression to predict the intersection-over-union (IoU) score without using any ground-truth labels during deployment. These approaches rely on the object detection output and hand-crafted features to evaluate object detection quality in run-time. Rabiee *et al*. [54] introduced a framework named introspective vision for obstacle avoidance (iVOA) consisting of a perception system and an introspection module for the task of obstacle avoidance. The introspection module is trained to detect false-positive and false-negative patches of input images where the perception system fails to detect obstacles. The authors demonstrated the feasibility of the proposed introspection model for both indoor and outdoor dataset.

### 2.1.2 Experiences in the Same Workspace or Context

As mobile robots often operate in the same places or the same contexts over long periods encountering periodic and seasonal variations in the deployment conditions, performance monitoring and failure prediction methods can take advantage of this knowledge. One example is the work by Hawke *et al*. [29] where they introduced an Experience-Based Classification (EBC) framework to improve mobile robot performance for pedestrian detection. They applied multiple scene filters to identify false-positive errors made by the pedestrian detector. They used those filtered out images to re-train the detector to achieve better performance on the same location during the next traversal. Through experimental evaluation, EBC was shown to be a viable alternative to hard-negative-mining without manually labelled data.

In the context of mobile robot Teach and Repeat, [9] proposed a *localization envelop* to capture the likely localization performance from the Teach phase to improve the performance during the Repeat phase. However, this approach is location-dependent and requires multiple Teach phases to learn the expected performance. To improve upon this work, Dequaire *et al*. [16] proposed an appearance-based approach to predict the localization envelop using a single Teach pass.

Using a probabilistic framework, Gurau *et al*. [26] predicted perception performance of a pedestrian detection system deployed on a mobile robot based on its previous visits to the same location. They estimated the detection performance for a particular place and granted or denied autonomy to the mobile robot based on the predicted performance. Most recently, in the Simultaneous Localization and Mapping (SLAM) paradigm, Rabiee *et al*. [53] proposed the idea of introspective vision-based SLAM. A self-supervised approach for learning to predict sources of failure for visual SLAM and to estimate a context-aware noise model for image correspondences, moving objects, non-rigid objects and other causes of errors.

In the context of Autonomous Vehicles (AV), Hecker *et al*. [30] argued that failure in the onboard vision system is not uncommon, and this does not happen randomly. Heavy traffic, complex intersections, adverse weather and illumination condition are conditions where the vision system will fail. They presented a method to learn to predict how challenging an environment is to a given vision-based model. Their proposed work predicts whether the current driving conditions are safe or hazardous for an underlying speed and steering angle prediction network that uses images collected from a vehicle's front-view camera.

### 2.2 Monitoring based on inconsistencies during inference

Another trend we can identify in the performance monitoring literature is the detection of inconsistency during the inference period. This inconsistency detection can be an indicator of performance degradation. The proposed approaches focus on using temporal and stereo vision, multiple sensor modalities, misalignment detection between the input and the output and abnormal neuron activation pattern. This section will provide a brief overview of these approaches.

Ramanagopal *et al*. [58] proposed using stereo and temporal inconsistency of a deployed object detection system to identify false negative instances. The stereo disparity is used to transfer detected object from one camera view to another for stereo inconsistency detection. A multi-object tracker is used to construct tracklets using the detected objects, and any missing tracklets in subsequent frames work as a false negative hypothesis.

Building on the literature of multiprocessor diagnosability, Antonante *et al*. [2] developed the temporal diagnostic graphs, a framework to reason over the consistency of perception outputs over time and demonstrated the ability to detect perception failures in an autonomous driving simulator.

Zhou *et al*. [78] proposes an automatic validation pipeline incorporating an additional sensor (LIDAR) to examine the performance of a semantic segmentation model in run-time. Using the geometric properties of neighbouring LIDAR points, they recognized road boundaries near the vehicle and automatically generated labels data for the road. By comparing the road segmentation model's predictions with the automatically generated labels, they measured the segmentation models accuracy at run-time.

Yang *et al*. [75] introduced the concept of mirrorability and mirror-error for object part localization, and showed that mirror-error could be measured without any ground-truth data. They also showed a high correlation between the mirror-error and the corresponding ground-truth error. Because of this correlation, mirror-error can be used to indicate localization/alignment error at run-time.

Gupta *et al*. [25] proposed an Adversarially-Trained Online Monitor (ATOM) to track the performance of neural networks that estimate 3D human shapes and poses from images. They address this problem by identifying the alignment inconsistency between the input image and the output mesh of a human shape and pose reconstruction network, GraphCMR [40]. ATOM generates a mesh correctness score and uses that to monitor the performance of GraphCMR prediction.

Henzinger *et al*. [34] proposed an abstraction-based framework to monitor a neural network by observing its hidden layers. This framework is a neural network architecture-independent, and the proposed abstraction represents all values encountered in the chosen layers during the training phase. During deployment, run-time monitoring is performed by comparing the current values in the layers with the abstraction. Another related work is proposed by Cheng *et al*. [6]. They stored the neuron activation pattern in an abstract form and used Hamming distance to compare the generated pattern at run-time to the abstract form. This comparison detects whether the run-time prediction made by the network is consistent with the prior training data.

## 2.3 Monitoring based on uncertainty estimation and confidence

In this section, we will provide a brief overview of the literature that focus to monitor the performance of a perception system using uncertainty estimation, prediction confidence and quality scores.

### 2.3.1 Uncertainty Estimation

Uncertainty estimation is an active area of deep learning research. It includes approaches as simple as softmax entropy [32] to more principled methods such Bayesian Neural Networks [44] and their approximation [21], and ensemble techniques [41]. For a comprehensive review of uncertainty estimation methods used in machine learning and deep learning see [1]. Due to the promising role of uncertainty estimation in increasing autonomous and robotic systems' safety by indicating low confidence in output predictions – and consequently detecting failures – many authors in the field of robotic perception investigated and compared variations of the main methods of estimating uncertainty from DNNs. Examples include uncertainty estimation for steering angle estimation [36], road segmentation [51], visual odometry [14], and vehicle and object detection [17, 47, 28].

The work by Grimmett *et al*. [24] is one of the earlier attempts to use uncertainty to monitor learning-based robotic perception. They showed that, in the robotic context, traditional performance metrics are inadequate to train and evaluate classifiers used for mission-critical decision making. To overcome this shortcoming, they proposed the concept of introspection – the ability to assess confidence to mitigate overconfident classifications. Based on this idea, they analyzed the introspective capability expressed using uncertainty estimation of multiple image classifiers and suggested using model ensemble instead of using a single model to take critical decisions in safety-critical robotic applications.

In the context of end-to-end controllers for self-driving cars, Michelmore *et al*. [46] explored the effectiveness of multiple measures of uncertainty and showed that mutual information, a measure of epistemic uncertainty [20], is a promising indicator of forthcoming crashes of the car. The evaluation was done using self-driving car simulator. In the context of vehicle detection, Feng *et al*. [18] proposed a probabilistic LIDAR vehicle detection network that captures model epistemic uncertainty by Monte Carlo Dropout [22] and aleatoric uncertainty [39] by adding an auxiliary output layer to the vehicle detection network.

Tian *et al*. [72] showed that different uncertainty measures correlate differently to different types of sensory data degradation, and proposed a method to combine multiple types of uncertainties in an adaptive fusion scheme for unseen degradation with application to RGB-D semantic segmentation.

Henne *et al*. [33] compared several methods for estimating uncertainty for image classification task against safety-related requirements and metrics designed to measure the model's performance in safety-critical domains. Their findings emphasize the repeatedly reported observation that Deep Ensembles [41] method for estimating uncertainty demonstrates strong performance. They also found that learned-confidence methods, the subject of the next section, produce consistently low confidence scores and can reject false predictions while producing higher confidence scores for correct predictions.

### 2.3.2 Confidence and Quality Scores

As shown in [50], deep neural networks often produce erroneous predictions with high confidence (low predictive uncertainty) when tested with data that differ from their training and test set. This is frequently the case for DNNs deployed on mobile robots in open-world settings. An emerging research trend for failure prediction is learning a specialized confidence score that acts as a measure for the quality of the target model outputs or as an indicator of the difficulty of the input to flag potential low-quality predictions.

For estimating model confidence, Corbiere *et al*. [11] defined a new confidence criterion called the True Class Probability (TCP) and proposed a network, ConfidNet, to learn the target confidence criterion. They provided theoretical guarantees and empirical evidence that predicting TCP instead of using maximum class probability (MCP) directly is better at predicting the failure of convolutional neural networks for a classification and a semantic segmentation task.

An example of the use of quality score is the works by Rottman *et al*. [62]. They proposed a meta-classifier to monitor the performance of a semantic segmentation model. The proposed approach uses pixel-wise uncertainty estimation and hand-crafted features corresponding to the target model segmentation's geometry to train a meta classifier or regressor to predict the IoU score with unknown ground-truth at run-time. Maag *et al*. [43] extend the work to account for temporal dependency between the input frames. As for input hardness prediction, Wang *et al*. [74] proposed an adversarially trained hardness predictor for a convolutional neural network classifier. The hardness-predictor is an auxiliary network that predicts a score for each input to the classifier denoting how hard it will be on the classifier. Based on this score, the classifier can either accept to classify the image or reject it altogether.

Although not directly applied to a robotic application, the approach of Valindria *et al*. [73] to semantic segmentation quality monitoring can be extended to robotic perception. They introduced the concept of Reverse Classification Accuracy (RCA) to evaluate a deployed segmentation model's performance without using any ground-truth labels. RCA is a reverse classifier trained using the predictions of the target segmentation model as pseudo-ground-truth. Dice similarity coefficient (DSC) – aka F1-score – between RCA's outputs and the target model's predictions is used as a quality score. Instead of applying RCA to predict the DSC, Robinson *et al*. [61] proposed to use a convolutional neural network. Their approach provides real-time inference and better accuracy for predicting the DSC for image segmentation task.

### 2.3.3 Out-of-Distribution Detection

Throughout the literature, out-of-distribution (OOD) detection is referred by multiple terms, for example anomaly, novelty or outlier detection [63, 45]. Nevertheless, these approaches' common objective is to identify testing samples that do not belong to the training set's data distribution. Concretely, let us assume we have trained a perception system to perform some specific task – image classification, segmentation or object detection, using a dataset sampled from the distribution $D_{in}$. Any dataset that is not a member of $D_{in}$ will be referred to as out-of-distribution. At run-time, we want to detect when the input comes from a distribution very different from $D_{in}$. (See [67] for a recent review of the different methods that tackle OOD detection and [52] for an empirical evaluation of several of these methods). In the context of robots that operate in open-world settings, this knowledge is essential since the DNNs could make an over-confidently wrong predictions when operating on out-of-distribution data. Upon identifying out-of-distribution input, the mobile robot can enable a fail-safe mode or hand-over control to a human operator's to ensure safety and reliability. This section discusses examples that use ideas related to OOD detection to monitor mobile robot performance.

In the context of a mobile robot's safe visual navigation, Richter *et al*. [60] proposed using an autoencoder along with a collision-avoidance system. The autoencoder decides whether an input image is similar enough to the training data to be confident about the collision avoidance system's prediction. In the case of low confidence, the mobile robot reverts to safe recovery behaviour which reduced the number of collisions and resulted in faster navigation time than the baseline approach. Cai *et al*. [3] demonstrated the application of out-of-distribution control input detection in the context of a self-driving car. Their approach is based on conformal prediction [68] and anomaly detection. The nonconformity score is computed using a variational autoencoder and a deep support vector machine. The experimental results show a decrease in the number of false-positive errors and a faster execution time during inference.

Nitsch *et al*. [49] proposed an uncertainty-based OOD detection technique that uses auxiliary training along with post-hoc statistics without requiring any external out-of-distribution dataset. The proposed approach takes advantage of Generative Adversarial Network (GAN) to enforce the object classifier to assign low confidence on OOD data and uses cosine similarity to identify OOD samples. Whereas Che *et al*. [4] proposed Deep Verifier Network to detect OOD and adversarial input to a deep neural network using conditional variational autoencoder.

Recently, Jafarzadeh *et al*. [37] formalized the open-world recognition reliability problem and proposed multiple automatic reliability assessment policies using only the reported probability distribution of a classifier. The proposed open-world reliability assessment works for both closed-set and open-set settings and shows significant improvement over a baseline algorithm.

A related topic to the approaches in this category is abstention or rejection learning, which is concerned with designing robust model that can reject an input assuming the possibility of making a wrong decision. Abstention learning can be used as an implicit approach for run-time performance monitoring. In abstention learning, each error and rejection incur a predefined cost, and the goal of abstention learning is to keep this error-reject cost at an optimal level. The error-reject tradeoff was first introduced by Chow [8, 7], where the author formalized the optimal rejection rule and derived the relation between the error and rejection probabilities. Following this work, [19], [31] and [12] introduced a rejection option to Support Vector Machines, nearest neighbours and boosting algorithms respectively. The rejection module in these approaches is trained separately from the targeted perception approach. Later Cortes *et al*. [13] and Geifman *et al*. [23] proposed rejection option that can be jointly learned with the perception system. [23] integrated a reject option with a deep neural network.

## 3 Run-time monitoring mapped to the robotic perception pipeline

Another way to categorize the papers reviewed in this survey is by where they perform the monitoring in the robotic perception pipeline (Figure 1). We can categorize the literature above according to whether they perform the monitoring by input validation or output evaluation or inner activations inspection or a combination of them.

Input validation means the performance monitoring system directly uses the same input as the perception system to predict failures and/or monitor the performance. As an example, [76] and [15] predict the success or failure of the perception system using a classifier that uses the same input as the perception system. In this case, performance monitoring is separated from the perception system. Output evaluation refers to the cases where performance monitoring is done by evaluating the perception system's output to predict its quality and express low or high confidence in it. [58] and [62] are examples of this paradigm. The activation inspection related research utilizes the perception system DNN's internal layer activations to monitor its performance and detect failures. [34] and [11] are examples of this category.

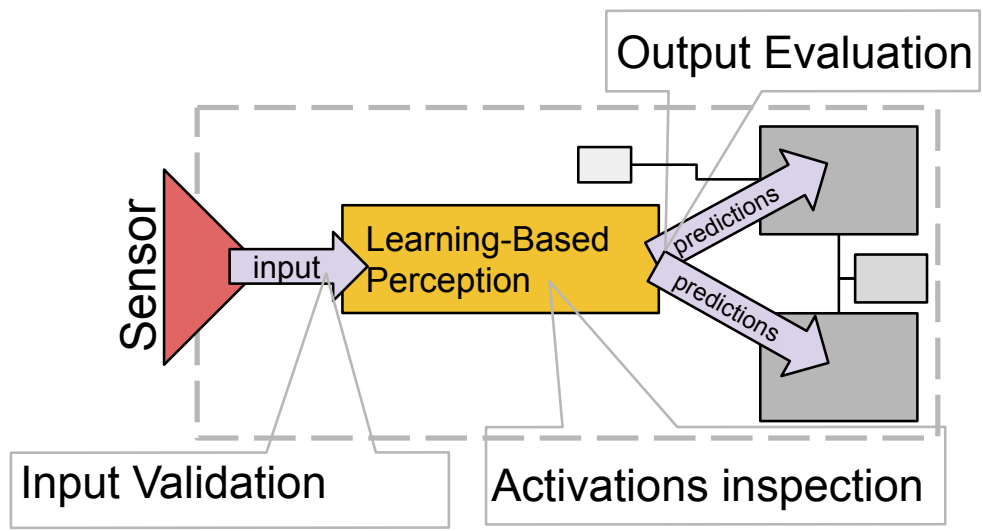

Figure 1: In a robotic system that uses a learning-based perception module, we can categorize the methods reviewed in this survey under methods that perform the monitoring by input validation or output evaluation or inner activations inspection or a combination of the above.

Table 1: Re-categorization of the papers based on where in the robotic perception pipeline they perform the monitoring and whether they required training during design-time or not.

| Paper | Input Validation | Activation Inspection | Output Evaluation | Explicit | Implicit |
|---|---|---|---|---|---|
| Jammalamadaka et al. [38] | ✓ | | | ✓ | |
| Zhang et al. [76] | ✓ | | | ✓ | |
| Daftry et al. [15] | ✓ | | | ✓ | |
| Saxena et al. [65] | ✓ | | | ✓ | |
| Mohseni et al. [48] | | ✓ | | ✓ | |
| Shao et al. [69] | | | ✓ | ✓ | |
| Rahman et al. [56, 57] | | ✓ | | ✓ | |
| Rahman et al. [55] | | ✓ | ✓ | | ✓ |
| Schubert et al. [66] | | | ✓ | ✓ | |
| Rabiee et al. [54] | ✓ | | | ✓ | |
| Valindria et al. [73] | | | ✓ | | ✓ |
| Robinson et al. [61] | | | ✓ | ✓ | |
| Cortes et al. [13] | ✓ | | | | ✓ |
| Geifman et al. [23] | ✓ | | | | ✓ |
| Hawke et al. [29] | | | ✓ | | ✓ |
| Churchill et al. [9] | ✓ | | | | ✓ |
| Gurau et al. [26] | ✓ | | | ✓ | |
| Rabiee et al. [53] | ✓ | | ✓ | ✓ | |
| Hecker et al. [30] | ✓ | | | ✓ | |
| Ramanagopal et al. [58] | | | ✓ | | ✓ |
| Yang et al. [75] | | | ✓ | | ✓ |
| Gupta et al. [25] | ✓ | | ✓ | ✓ | |
| Henzinger et al. [34] | | ✓ | | | ✓ |
| Antonante et al. [2] | | | ✓ | | ✓ |
| Grimmett et al. [24] | | | ✓ | | ✓ |
| Feng et al. [18] | | | ✓ | | ✓ |
| Corbiere et al. [11] | | ✓ | | | ✓ |
| Rottman et al. [62] | | | ✓ | ✓ | |
| Wang et al. [74] | ✓ | | | | ✓ |
| Richter et al. [60] | ✓ | | | | ✓ |
| Cai et al. [3] | ✓ | | | | ✓ |
| Tian et al. [72] | | | ✓ | | ✓ |
| Nitsch et al. [49] | | | ✓ | | ✓ |
| Che et al. [4] | | | ✓ | | ✓ |
| Jafarzadeh et al. [37] | | | ✓ | | ✓ |

Moreover, we can categorize the performance monitoring literature into *explicit* and *implicit* monitoring. In *explicit* monitoring, performance monitoring utilizes examples of success and failure from design-time before deployment. As an example, in [69] and [48], the performance monitor is trained using the perception system input and their corresponding accuracy and steering error, respectively. On the other hand, *implicit* monitoring does not require training using examples of success or failure. For example, [58] uses the stereo and temporal inconsistency to identify false-negatives and [60] identifies inputs dissimilar to the training set as a potential cause of navigation failure. Table 1 lists all the papers and their corresponding categorization.

## 4 Conclusion

Run-time monitoring of learning-based perception systems – dominated by deep neural networks – is crucial for robotic applications due to the difficulty in applying design-time formal verification and safety guarantees for such systems, mainly due to their complexity and the complexity of modelling their deployment environments. In this survey, we identified an emerging research direction that focuses on run-time verification and monitoring. The approaches we reviewed tackle the problem in various ways. Some depend on past experiences and examples of success and failures to train a monitoring system that verifies some input/output/neural activations properties for the target model. Other approaches detected run-time inconsistencies in the input/output/internal activations as a mean to predict failures. The last group of methods use uncertainty estimation, learned confidence, and detect out-of-distribution input to predict the low-quality output from the target model. We also mapped these approaches based on where they perform the monitoring in the perception system pipeline and whether they require training during design-time. Due to the importance of this line of research for many safety-critical systems that use learning-based components such as deep neural networks with millions of parameters, a more principled approach to run-time monitoring is needed – one that considers not only the target perception module by itself but also the whole robotic system and the interaction between its various modules overtime.

# References

[1] Moloud Abdar, Farhad Pourpanah, Sadiq Hussain, Dana Rezazadegan, Li Liu, Mohammad Ghavamzadeh, Paul Fieguth, Abbas Khosravi, U Rajendra Acharya, Vladimir Makarenkov, et al. A Review of Uncertainty Quantification in Deep Learning: Techniques, Applications and Challenges. *arXiv preprint arXiv:2011.06225*, 2020.

[2] Pasquale Antonante, David I. Spivak, and L. Carlone. Monitoring and Diagnosability of Perception Systems. *ArXiv*, abs/2011.07010, 2020.

[3] Feiyang Cai and Xenofon Koutsoukos. Real-Time Out-Of-Distribution Detection in Learning-Enabled Cyber-Physical Systems. In *2020 ACM/IEEE 11th International Conference on Cyber-Physical Systems (ICCPS)*, pages 174–183. IEEE, 2020.

[4] Tong Che, Xiaofeng Liu, Site Li, Yubin Ge, Ruixiang Zhang, Caiming Xiong, and Yoshua Bengio. Deep Verifier Networks: Verification of Deep Discriminative Models with Deep Generative Models. *arXiv preprint arXiv:1911.07421*, 2019.

[5] Liang-Chieh Chen, G. Papandreou, I. Kokkinos, Kevin Murphy, and A. Yuille. DeepLab: Semantic Image Segmentation with Deep Convolutional Nets, Atrous Convolution, and Fully Connected CRFs. *IEEE Transactions on Pattern Analysis and Machine Intelligence*, 40:834–848, 2018.

[6] Chih-Hong Cheng, Georg Nührenberg, and Hirotoshi Yasuoka. Runtime Monitoring Neuron Activation Patterns. *2019 Design, Automation Test in Europe Conference Exhibition (DATE)*, pages 300–303, 2019.

[7] C Chow. On Optimum Recognition Error and Reject Tradeoff. *IEEE Transactions on information theory*, 16(1):41–46, 1970.

[8] Chi-Keung Chow. An Optimum Character Recognition System using Decision Functions. *IRE Transactions on Electronic Computers*, (4):247–254, 1957.

[9] W. Churchill, C. Tong, C. Gurau, I. Posner, and P. Newman. Know Your Limits: Embedding Localiser Performance Models in Teach and Repeat Maps. *2015 IEEE International Conference on Robotics and Automation (ICRA)*, pages 4238–4244, 2015.

[10] Felipe Codevilla, M. Müller, A. Dosovitskiy, Antonio López, and V. Koltun. End-To-End Driving via Conditional Imitation Learning. *2018 IEEE International Conference on Robotics and Automation (ICRA)*, pages 1–9, 2018.

[11] Charles Corbière, Nicolas Thome, Avner Bar-Hen, Matthieu Cord, and Patrick Pérez. Addressing Failure Prediction by Learning Model Confidence. *Advances in Neural Information Processing Systems*, 32:2902–2913, 2019.

[12] Corinna Cortes, Giulia DeSalvo, and Mehryar Mohri. Boosting with Abstention. *Advances in Neural Information Processing Systems*, 29:1660–1668, 2016.

[13] Corinna Cortes, Giulia DeSalvo, and Mehryar Mohri. Learning with Rejection. In *International Conference on Algorithmic Learning Theory*, pages 67–82. Springer, 2016.

[14] G. Costante and M. Mancini. Uncertainty Estimation for Data-Driven Visual Odometry. *IEEE Transactions on Robotics*, 36:1738–1757, 2020.

[15] Shreyansh Daftry, Sam Zeng, J Andrew Bagnell, and Martial Hebert. Introspective Perception: Learning to Predict Failures in Vision Systems. In *2016 IEEE/RSJ International Conference on Intelligent Robots and Systems (IROS)*, pages 1743–1750. IEEE, 2016.

[16] J. Dequaire, C. Tong, W. Churchill, and I. Posner. Off the Beaten Track: Predicting Localisation Performance in Visual Teach and Repeat. *2016 IEEE International Conference on Robotics and Automation (ICRA)*, pages 795–800, 2016.

[17] Di Feng, Ali Harakeh, Steven L. Waslander, and K. Dietmayer. A Review and Comparative Study on Probabilistic Object Detection in Autonomous Driving. *ArXiv*, abs/2011.10671, 2020.

[18] Di Feng, L. Rosenbaum, and K. Dietmayer. Towards Safe Autonomous Driving: Capture Uncertainty in the Deep Neural Network for Lidar 3D Vehicle Detection. *2018 21st International Conference on Intelligent Transportation Systems (ITSC)*, pages 3266–3273, 2018.

[19] Giorgio Fumera and Fabio Roli. Support Vector Machines with Embedded Reject Option. In *International Workshop on Support Vector Machines*, pages 68–82. Springer, 2002.

[20] Yarin Gal. Uncertainty in Deep Learning. *University of Cambridge*, 1(3), 2016.

[21] Yarin Gal and Zoubin Ghahramani. Dropout as a bayesian approximation: Representing model uncertainty in deep learning. In *Proceedings of the 33rd International Conference on International Conference on Machine Learning - Volume 48*, ICML'16, page 1050–1059. JMLR.org, 2016.

[22] Yarin Gal, Jiri Hron, and Alex Kendall. Concrete Dropout. In *Advances in neural information processing systems*, pages 3581–3590, 2017.

[23] Yonatan Geifman and Ran El-Yaniv. Selective Classification for Deep Neural Networks. In *Advances in neural information processing systems*, pages 4878–4887, 2017.

[24] Hugo Grimmett, Rudolph Triebel, R. Paul, and I. Posner. Introspective Classification for Robot Perception. *The International Journal of Robotics Research*, 35:743 – 762, 2016.

[25] Arjun Gupta and L. Carlone. Online Monitoring for Neural Network Based Monocular Pedestrian Pose Estimation. *ArXiv*, abs/2005.05451, 2020.

[26] C. Gurau, D. Rao, C. Tong, and I. Posner. Learn from Experience: Probabilistic Prediction of Perception Performance to Avoid Failure. *The International Journal of Robotics Research*, 37:981 – 995, 2018.

[27] D. Hall, Feras Dayoub, T. Perez, and C. McCool. A Rapidly Deployable Classification System using Visual Data for the Application of Precision Weed Management. *Comput. Electron. Agric.*, 148:107–120, 2018.

[28] Ali Harakeh, M. Smart, and Steven L. Waslander. BayesOD: A Bayesian Approach for Uncertainty Estimation in Deep Object Detectors. *2020 IEEE International Conference on Robotics and Automation (ICRA)*, pages 87–93, 2020.

[29] J. Hawke, C. Gurau, C. Tong, and I. Posner. Wrong Today, Right Tomorrow: Experience-Based Classification for Robot Perception. In *FSR*, 2015.

[30] S. Hecker, Dengxin Dai, and L. Gool. Failure Prediction for Autonomous Driving. *2018 IEEE Intelligent Vehicles Symposium (IV)*, pages 1792–1799, 2018.

[31] Martin E Hellman. The Nearest Neighbor Classification Rule with a Reject Option. *IEEE Transactions on Systems Science and Cybernetics*, 6(3):179–185, 1970.

[32] Dan Hendrycks and Kevin Gimpel. A Baseline for Detecting Misclassified and Out-of-Distribution Examples in Neural Networks. *Proceedings of International Conference on Learning Representations*, 2017.

[33] Maximilian Henne, Adrian Schwaiger, K. Roscher, and G. Weiss. Benchmarking Uncertainty Estimation Methods for Deep Learning with Safety-Related Metrics. In *SafeAI@AAAI*, 2020.

[34] T. Henzinger, Anna Lukina, and C. Schilling. Outside the Box: Abstraction-Based Monitoring of Neural Networks. *ArXiv*, abs/1911.09032, 2020.

[35] Xiaowei Huang, Daniel Kroening, Wenjie Ruan, James Sharp, Youcheng Sun, Emese Thamo, Min Wu, and Xinping Yi. A Survey of Safety and Trustworthiness of Deep Neural Networks: Verification, Testing, Adversarial Attack and Defence, and Interpretability. *Computer Science Review*, 37:100270, 2020.

[36] Christian Hubschneider, Robin Hutmacher, and Johann Marius Zöllner. Calibrating Uncertainty Models for Steering Angle Estimation. *2019 IEEE Intelligent Transportation Systems Conference (ITSC)*, pages 1511–1518, 2019.

[37] Mohsen Jafarzadeh, Touqeer Ahmad, Akshay Raj Dhamija, Chunchun Li, Steve Cruz, and Terrance E Boult. Automatic Open-World Reliability Assessment. *arXiv preprint arXiv:2011.05506*, 2020.

[38] Nataraj Jammalamadaka, Andrew Zisserman, Marcin Eichner, Vittorio Ferrari, and CV Jawahar. Has My Algorithm Succeeded? An Evaluator for Human Pose Estimators. In *European Conference on Computer Vision*, pages 114–128. Springer, 2012.

[39] Alex Kendall and Yarin Gal. What Uncertainties Do We Need in Bayesian Deep Learning for Computer Vision? In *Advances in neural information processing systems*, pages 5574–5584, 2017.

[40] Nikos Kolotouros, Georgios Pavlakos, and Kostas Daniilidis. Convolutional Mesh Regression for Single-Image Human Shape Reconstruction. In *Proceedings of the IEEE Conference on Computer Vision and Pattern Recognition*, pages 4501–4510, 2019.

[41] Balaji Lakshminarayanan, A. Pritzel, and Charles Blundell. Simple and Scalable Predictive Uncertainty Estimation using Deep Ensembles. In *NIPS*, 2017.

[42] Li Liu, Wanli Ouyang, X. Wang, P. Fieguth, J. Chen, Xinwang Liu, and M. Pietikäinen. Deep Learning for Generic Object Detection: A Survey. *International Journal of Computer Vision*, 128:261–318, 2019.

[43] Kira Maag, M. Rottmann, and H. Gottschalk. Time-Dynamic Estimates of the Reliability of Deep Semantic Segmentation Networks. *2020 IEEE 32nd International Conference on Tools with Artificial Intelligence (ICTAI)*, pages 502–509, 2020.

[44] D. MacKay. A Practical Bayesian Framework for Backpropagation Networks. *Neural Computation*, 4:448–472, 1992.

[45] Marc Masana, Idoia Ruiz, Joan Serrat, Joost van de Weijer, and Antonio M Lopez. Metric Learning for Novelty and Anomaly Detection. *arXiv preprint arXiv:1808.05492*, 2018.

[46] Rhiannon Michelmore, M. Kwiatkowska, and Yarin Gal. Evaluating Uncertainty Quantification in End-To-End Autonomous Driving Control. *ArXiv*, abs/1811.06817, 2018.

[47] Dimity Miller, Feras Dayoub, Michael Milford, and Niko Sünderhauf. Evaluating Merging Strategies for Sampling-Based Uncertainty Techniques in Object Detection. In *2019 International Conference on Robotics and Automation (ICRA)*, pages 2348–2354. IEEE, 2019.

[48] Sina Mohseni, Akshay Jagadeesh, and Zhangyang Wang. Predicting Model Failure using Saliency Maps in Autonomous Driving Systems. *arXiv preprint arXiv:1905.07679*, 2019.

[49] Julia Nitsch, Masha Itkina, Ransalu Senanayake, Juan Nieto, Max Schmidt, Roland Siegwart, Mykel J Kochenderfer, and Cesar Cadena. Out-Of-Distribution Detection for Automotive Perception. *arXiv preprint arXiv:2011.01413*, 2020.

[50] Yaniv Ovadia, Emily Fertig, Jie Ren, Zachary Nado, David Sculley, Sebastian Nowozin, Joshua Dillon, Balaji Lakshminarayanan, and Jasper Snoek. Can You Trust Your Model's Uncertainty? Evaluating Predictive Uncertainty Under Dataset Shift. In *Advances in Neural Information Processing Systems*, pages 13991–14002, 2019.

[51] Buu Phan, S. Khan, Rick Salay, and K. Czarnecki. Bayesian Uncertainty Quantification with Synthetic Data. In *SAFECOMP Workshops*, 2019.

[52] Stephan Rabanser, Stephan Günnemann, and Zachary Chase Lipton. Failing Loudly: An Empirical Study of Methods for Detecting Dataset Shift. *ArXiv*, abs/1810.11953, 2019.

[53] S. Rabiee and J. Biswas. IV-SLAM: Introspective Vision for Simultaneous Localization and Mapping. *ArXiv*, abs/2008.02760, 2020.

[54] Sadegh Rabiee and Joydeep Biswas. IVOA: Introspective Vision for Obstacle Avoidance. *arXiv preprint arXiv:1903.01028*, 2019.

[55] Quazi Marufur Rahman, Niko Sünderhauf, and Feras Dayoub. Did You Miss the Sign? A False Negative Alarm System for Traffic Sign Detectors. *2019 IEEE/RSJ International Conference on Intelligent Robots and Systems (IROS)*, pages 3748–3753, 2019.

[56] Quazi Marufur Rahman, Niko Sünderhauf, and Feras Dayoub. Online Monitoring of Object Detection Performance Post-Deployment. *arXiv preprint arXiv:2011.07750*, 2020.

[57] Quazi Marufur Rahman, Niko Sünderhauf, and Feras Dayoub. Per-Frame mAP Prediction for Continuous Performance Monitoring of Object Detection During Deployment. In *Proceedings of the IEEE/CVF Winter Conference on Applications of Computer Vision (WACV) Workshops*, pages 152–160, January 2021.

[58] Manikandasriram Srinivasan Ramanagopal, Cyrus Anderson, R. Vasudevan, and M. Johnson-Roberson. Failing to Learn: Autonomously Identifying Perception Failures for Self-Driving Cars. *IEEE Robotics and Automation Letters*, 3:3860–3867, 2018.

[59] Benjamin Recht, Rebecca Roelofs, Ludwig Schmidt, and Vaishaal Shankar. Do ImageNet Classifiers Generalize to ImageNet? volume 97 of *Proceedings of Machine Learning Research*, pages 5389–5400, Long Beach, California, USA, 09–15 Jun 2019. PMLR.

[60] Charles Richter and Nicholas Roy. Safe Visual Navigation via Deep Learning and Novelty Detection. In *Robotics: Science and Systems XIII, Massachusetts Institute of Technology, Cambridge, Massachusetts, USA, July 12-16*, 2017.

[61] R. Robinson, O. Oktay, Wenjia Bai, V. Valindria, Mihir M. Sanghvi, Nay Aung, J. Paiva, F. Zemrak, K. Fung, E. Lukaschuk, A. Lee, Valentina Carapella, Y. Kim, Bernhard Kainz, S. Piechnik, S. Neubauer, S. Petersen, Chris Page, D. Rueckert, and Ben Glocker. Real-Time Prediction of Segmentation Quality. In *MICCAI*, 2018.

[62] M. Rottmann, P. Colling, Thomas-Paul Hack, Fabian Hüger, Peter Schlicht, and H. Gottschalk. Prediction Error Meta Classification in Semantic Segmentation: Detection via Aggregated Dispersion Measures of Softmax Probabilities. *2020 International Joint Conference on Neural Networks (IJCNN)*, pages 1–9, 2020.

[63] Lukas Ruff, Jacob R Kauffmann, Robert A Vandermeulen, Grégoire Montavon, Wojciech Samek, Marius Kloft, Thomas G Dietterich, and Klaus-Robert Müller. A Unifying Review of Deep and Shallow Anomaly Detection. *arXiv preprint arXiv:2009.11732*, 2020.

[64] I. Sa, C. Lehnert, A. English, C. McCool, Feras Dayoub, B. Upcroft, and T. Perez. Peduncle Detection of Sweet Pepper for Autonomous Crop Harvesting—Combined Color and 3-D Information. *IEEE Robotics and Automation Letters*, 2:765–772, 2017.

[65] Dhruv Mauria Saxena, Vince Kurtz, and Martial Hebert. Learning Robust Failure Response for Autonomous Vision Based Flight. In *2017 IEEE International Conference on Robotics and Automation (ICRA)*, pages 5824–5829. IEEE, 2017.

[66] M. Schubert, K. Kahl, and M. Rottmann. MetaDetect: Uncertainty Quantification and Prediction Quality Estimates for Object Detection. *ArXiv*, abs/2010.01695, 2020.

[67] Alireza Shafaei, M. Schmidt, and J. Little. Does Your Model Know the Digit 6 Is Not a Cat? A Less Biased Evaluation of "Outlier" Detectors. *ArXiv*, abs/1809.04729, 2018.

[68] Glenn Shafer and Vladimir Vovk. A Tutorial on Conformal Prediction. *Journal of Machine Learning Research*, 9(Mar):371–421, 2008.

[69] Zhihui Shao, Jianyi Yang, and Shaolei Ren. Increasing Trustworthiness of Deep Neural Networks via Accuracy Monitoring. *Workshop on Artificial Intelligence Safety 2020*, 2020.

[70] Niko Sünderhauf, Feras Dayoub, S. McMahon, Ben Talbot, R. Schulz, P. Corke, G. Wyeth, B. Upcroft, and Michael Milford. Place Categorization and Semantic Mapping on a Mobile Robot. *2016 IEEE International Conference on Robotics and Automation (ICRA)*, pages 5729–5736, 2016.

[71] Mingxing Tan and Quoc V. Le. EfficientNet: Rethinking Model Scaling for Convolutional Neural Networks. In Kamalika Chaudhuri and Ruslan Salakhutdinov, editors, *Proceedings of the 36th International Conference on Machine Learning, ICML 2019, 9-15 June 2019, Long Beach, California, USA*, volume 97 of *Proceedings of Machine Learning Research*, pages 6105–6114. PMLR, 2019.

[72] Junjiao Tian, Wesley Cheung, Nathaniel Glaser, Yen-Cheng Liu, and Zsolt Kira. UNO: Uncertainty-Aware Noisy-Or Multimodal Fusion for Unanticipated Input Degradation. In *2020 IEEE International Conference on Robotics and Automation (ICRA)*, pages 5716–5723. IEEE, 2020.

[73] V. Valindria, I. Lavdas, Wenjia Bai, K. Kamnitsas, E. Aboagye, A. Rockall, D. Rueckert, and Ben Glocker. Reverse Classification Accuracy: Predicting Segmentation Performance in the Absence of Ground Truth. *IEEE Transactions on Medical Imaging*, 36:1597–1606, 2017.

[74] P. Wang and N. Vasconcelos. Towards Realistic Predictors. In *ECCV*, 2018.

[75] H. Yang and I. Patras. Mirror, mirror on the wall, tell me, is the error small? *2015 IEEE Conference on Computer Vision and Pattern Recognition (CVPR)*, pages 4685–4693, 2015.

[76] Peng Zhang, Jiuling Wang, Ali Farhadi, Martial Hebert, and Devi Parikh. Predicting Failures of Vision Systems. In *Proceedings of the IEEE Conference on Computer Vision and Pattern Recognition*, pages 3566–3573, 2014.

[77] Chaoqiang Zhao, Qiyu Sun, Chongzhen Zhang, Yang Tang, and Feng Qian. Monocular Depth Estimation Based on Deep Learning: An Overview. *Science China Technological Sciences*, pages 1–16, 2020.

[78] W. Zhou, J. Berrio, S. Worrall, and Eduardo M. Nebot. Automated Evaluation of Semantic Segmentation Robustness for Autonomous Driving. *IEEE Transactions on Intelligent Transportation Systems*, 21:1951–1963, 2020.